\documentclass[conference]{IEEEtran}
\IEEEoverridecommandlockouts

\usepackage{amsmath,amssymb,amsfonts}
\usepackage{algorithmic}
\usepackage{graphicx}
\usepackage{textcomp}
\usepackage{xcolor}
\usepackage{kotex}
\usepackage{multirow}
\usepackage{booktabs}
\usepackage[numbers]{natbib}
\usepackage{pgfplots}

\def\BibTeX{{\rm B\kern-.05em{\sc i\kern-.025em b}\kern-.08em
    T\kern-.1667em\lower.7ex\hbox{E}\kern-.125emX}}
\pgfplotsset{compat=1.18}
\begin{document}
\title{Learning Primitive Relations for\\Compositional Zero-Shot Learning 
\\

{\tiny \thanks{This work was supported by the National Research Foundation of Korea (NRF) and Institute of Information \& communications Technology Planning \& Evaluation (IITP) under the artificial intelligence semiconductor support program to nurture the best talents (IITP-2023-RS-2023-00256081) grant funded by the Korea government (MSIT)(RS-2023-00208985).
}}

}
\author{\IEEEauthorblockN{Insu Lee, Jiseob Kim, Kyuhong Shim and Byonghyo Shim}
\IEEEauthorblockA{Dept. of Electrical and Computer Engineering and Institute of New Media and Communications \\
Seoul National University, Seoul, Republic of Korea \\
\small{\texttt {  \{islee, jskim, khshim, bshim\}@islab.snu.ac.kr}}}
}

\maketitle

\begin{abstract}  
Compositional Zero-Shot Learning (CZSL) aims to identify unseen state-object compositions by leveraging knowledge learned from seen compositions.
Existing approaches often independently predict states and objects, overlooking their relationships.
In this paper, we propose a novel framework, learning primitive relations (LPR), designed to probabilistically capture the relationships between states and objects.
By employing the cross-attention mechanism, LPR considers the dependencies between states and objects, enabling the model to infer the likelihood of unseen compositions. 
Experimental results demonstrate that LPR outperforms state-of-the-art methods on all three CZSL benchmark datasets in both closed-world and open-world settings. 
Through qualitative analysis, we show that LPR leverages state-object relationships for unseen composition prediction.
\end{abstract}

\begin{IEEEkeywords}
Compositional Zero-Shot Learning, Vision-Language Model, Relation Learning
\end{IEEEkeywords}

\section{Introduction}
\label{sec:intro}
Although it sounds strange, when we hear the phrase \textit{purple cow}, we can anyway picture it in our minds.
The word \textit{purple cow} combines prior knowledge of state (`\textit{purple}') and object (`\textit{cow}').
In contrast to this, AI models often fail to classify new state-object compositions that have not been trained.
Recently, Compositional Zero-Shot Learning (CZSL)~\cite{misra2017fromredwinetoredtomato} has been introduced to train models such that it can classify unseen compositional classes without additional training.
Key idea of CZSL is that the model leverages knowledge about states and objects learned from seen compositions and then generalizes it to unseen compositions.
In CZSL, each class included in the states and objects is called \textit{primitive}.

\begin{figure}[t]
    \centering
    \includegraphics[width=0.85\columnwidth]{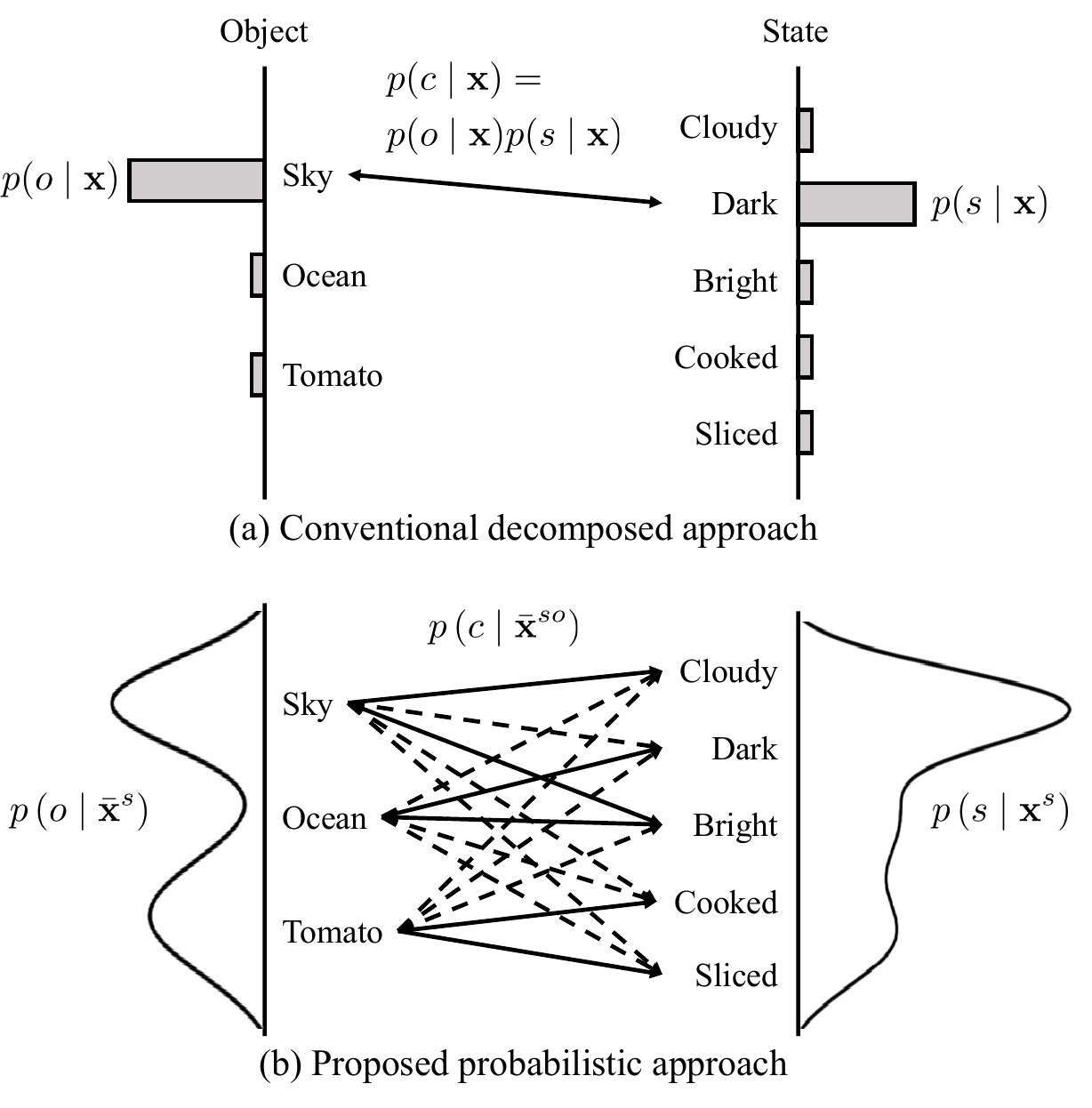}
    \caption{Illustration of the conventional and proposed CZSL approaches. While the conventional methods assume independence between states and objects, the proposed LPR learns probabilistic relationships between primitives.}
    \vspace{-0.5cm}
    \label{fig:intro}
\end{figure}

In recent studies~\cite{huang2024troika,li2024CDS-CZSL,wang2023HPL,jiang2024revealingLong-Tail_Distribution,lu2023DFSP}, generalization ability of pre-trained vision-language models (VLMs), such as CLIP~\cite{radford2021clip} is used to represent compositions in natural language.
In order to perform the classification properly, a feature of the ``purple cow" \textit{image} and that for the text \textit{``A photo of purple cow"} should be similar in the VLM feature space.
Previous works employ two separate branches that predict state and object independently.
These methods aim to decompose state and object information and then treat them as independent elements in an image (see Fig.~\ref{fig:intro}(a))~\cite{huang2024troika,li2024CDS-CZSL}.

While the abovementioned approaches are effective to some extent, they often fail to capture the relationships between states and objects.
We argue that understanding the relationships between them is essential for classifying unseen compositions.
The first reason is that the model can filter out nonsense compositions by learning the co-occurrence of similar objects or states.
For instance, when classifying the state of an unseen composition ``Dark Sky" (dashed lines in Fig.~\ref{fig:intro}(b)), the model is likely to assign low probabilities to states ``Cooked" or ``Sliced", since similar object ``Ocean" also exhibits low probabilities for these states.
The second reason is that the model can compute the likelihood of unseen compositions by learning the probabilistic relationships between states and objects.
For example, the probability of ``Dark Sky" can be modeled by referencing similar primitive relationships of seen compositions, such as ``Dark Ocean" and ``Bright Sky" (solid lines in Fig.~\ref{fig:intro}(b)).

\begin{figure*}[t]
    \centering
    \includegraphics[width=1.0\textwidth]{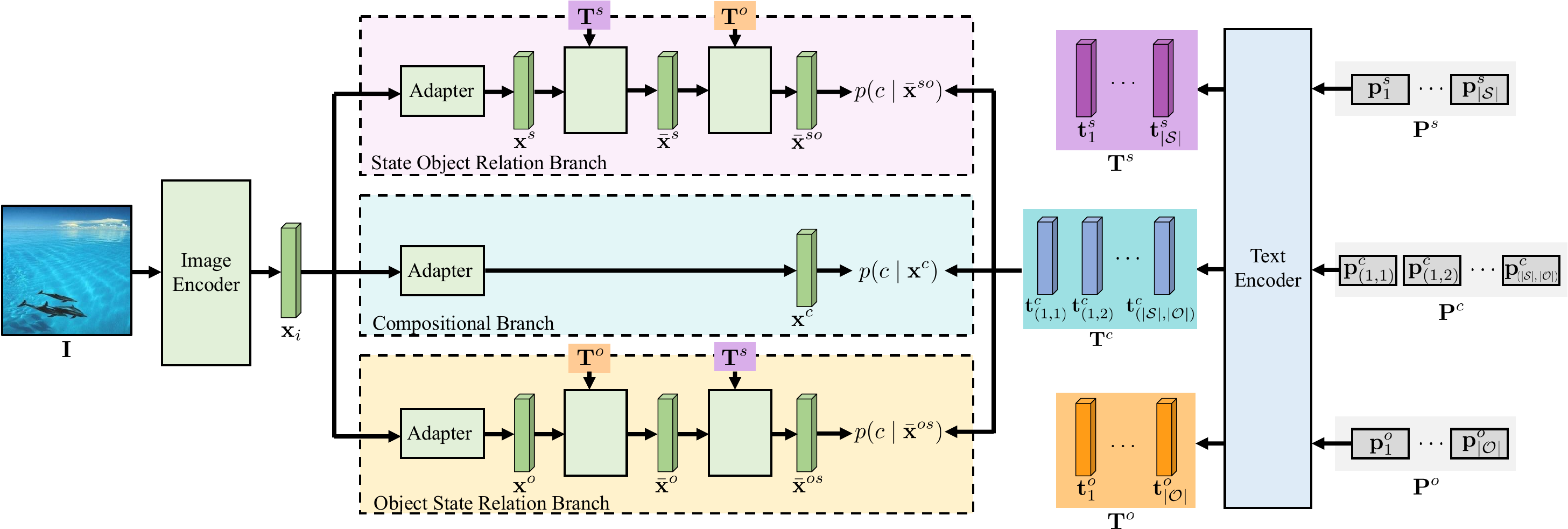}
    \caption{Overview of the proposed LPR. LPR consists of three branches, namely compositional (cyan, \textit{com}), state object relation (purple, \textit{sor}), and object state relation (yellow, \textit{osr}) branch. Best viewed in color.}
    \vspace{-0.3cm}
    \label{fig:main_fig}
\end{figure*}

In this paper, we propose a novel framework, called learning primitive relations (LPR), which captures relationships between primitives in a probabilistic manner.
Key idea of LPR is to extract object-related (or state-related) features conditioned on the state (or object) using cross-attention, which converts the similarity between images and primitives into probabilities.
Experimental results show that LPR achieves state-of-the-art (SOTA) performance in all three CZSL datasets.

\begin{figure}[t]
    \centering
    \includegraphics[width=1.0\columnwidth]{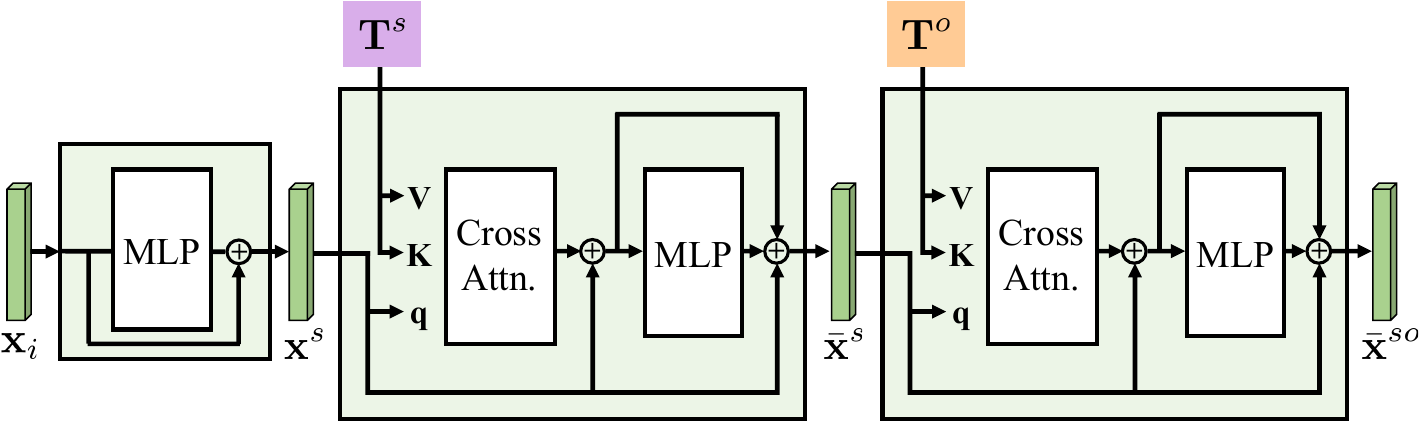}
    \caption{Architecture of the state object relation (\textit{sor}) branch.}
    \vspace{-0.3cm}
    \label{fig:main_sub}
\end{figure}

\section{Learning Primitive Relations}

\subsection{CZSL Formulation}
In CZSL, an image is classified into one of the classes within the compositional set  \( \mathcal{C} = \mathcal{S} \times \mathcal{O}  \), where \( \mathcal{S} \) represents a set of states  \( \{s_1, s_2, \dots, s_{|\mathcal{S}|}\} \) and \( \mathcal{O} \) represents a set of objects \( \{o_1, o_2, \dots, o_{|\mathcal{O}|}\} \).
During the training, the model can only access a set of seen compositions \(\mathcal{C}^{seen}\subset \mathcal{C}\).
The unseen compositions form the set $\mathcal{C}^{unseen}$, which is the subset of complement of $\mathcal{C}^{seen}$ (\( \mathcal{C}^{unseen} \subset \mathcal{C} \setminus \mathcal{C}^{seen} \)).
During the inference, the model is tested using both seen and unseen compositions.

\subsection{CLIP Feature Extraction for Compositional Classification}
In this work, we exploit the CLIP model~\cite{radford2021clip} to extract image and text features within a shared feature space.
The CLIP visual encoder processes an input image \(\mathbf{I}\) to extract the image feature $\mathbf{x}_i$.
Concurrently, the CLIP text encoder takes three types of prompts: $\mathbf{P}^s$, $\mathbf{P}^o$, and $\mathbf{P}^c$, which indicate state, object, and compositional prompts, respectively.
Then, the text encoder converts these prompts into text features $\mathbf{T}^s=[\mathbf{t}^{s}_1, \dots, \mathbf{t}^{s}_{|\mathcal{S}|}]$, $\mathbf{T}^o=[\mathbf{t}^{o}_1, \dots, \mathbf{t}^{o}_{|\mathcal{O}|}]$ and $\mathbf{T}^c=[\mathbf{t}^{c}_{(1,1)}, \dots, \mathbf{t}^{c}_{(|\mathcal{S}|,|\mathcal{O}|)}]$.
Note that the subscripts in the text features indicate the class indices.
For example, $\mathbf{t}^s_k$, $\mathbf{t}^o_l$ and $\mathbf{t}^c_{(k,l)}$ represent the $k$-th state, the $l$-th object, and the ($k,l$)-th compositional class, respectively~\cite{huang2024troika}.

\setlength{\tabcolsep}{8pt}
\begin{table*}[htbp]
    \centering
        \caption{Performance comparison under closed-world and open-world settings.  
        Metrics include seen accuracy (S), unseen accuracy (U), harmonic mean (HM), and area-under-curve (AUC).
        The best results are in \textbf{bold}. The second-best results are \underline{underlined}.
        }
        \vspace{-0.15cm}
    \renewcommand{\arraystretch}{1.1}
    \resizebox{1.0\textwidth}{!}{
    \begin{tabular}{c|c|c c c c|c c c c|c c c c}
        \hline
        \multirow{2}{*}{Setup} & \multirow{2}{*}{Method} & \multicolumn{4}{c|}{MIT-States} & \multicolumn{4}{c|}{UT-Zappos} & \multicolumn{4}{c}{C-GQA}  \\
        \cline{3-14}
        & & S & U & HM & AUC & S & U & HM & AUC & S & U & HM & AUC \\
        \hline
        \multirow{10}{*}{\rotatebox{90}{Closed-World}} 
        & CLIP~\cite{radford2021clip} \textsubscript{ICML'2021} & 30.2 & 46.0 & 26.1 & 11.0 & 15.8 & 49.1 & 15.6 & 5.0 & 7.5 & 25.0 & 8.6 & 1.4   \\
        & CoOp~\cite{zhou2022CoOp} \textsubscript{IJCV'2022} & 34.4 & 47.6 & 29.8 & 13.5 & 52.1 & 49.3 & 34.6 & 18.8 & 20.5 & 26.8 & 17.1 & 4.4   \\
        & Co-CGE~\cite{mancini2022Co-CGE} \textsubscript{TPAMI'2022} & 46.7 & 45.9 & 33.1 & 17.0 & 63.4 & 71.3 & 49.7 & 36.3 & 34.1 & 21.2 & 18.9 & 5.7  \\
        & CSP~\cite{nayak2022CSP} \textsubscript{ICLR'2023} & 46.6 & 49.9 & 36.3 & 19.4 & 64.2 & 66.2 & 46.6 & 33.0 & 28.8 & 26.8 & 20.5 & 6.2 \\
        & DFSP(t2i)~\cite{lu2023DFSP} \textsubscript{CVPR'2023} & 46.9 & 52.0 & 37.3 & 20.6 & 66.7 & 71.7 & 47.2 & 36.0 & 38.2 & 32.0 & 27.1 & 10.5 \\
        & GIPCOL~\cite{xu2024gipcol} \textsubscript{WACV'2024} & 48.5 & 49.6 & 36.6 & 19.9 & 65.0 & 68.5 & 48.8 & 36.2 & 31.9 & 28.4 & 22.5 & 7.1  \\
        & PLID~\cite{bao2023PLID} \textsubscript{ECCV'2024} & 49.7 & 52.4 & 39.0 & 22.1 & \textbf{67.3} & 68.8 & 52.4 & 38.7 & 38.8 & 33.0 & 27.9 & 11.0   \\
        & CDS-CZSL~\cite{li2024CDS-CZSL} \textsubscript{CVPR'2024}  &\underline{50.3} &52.9 &39.2 &\underline{22.4} &63.9 & \underline{74.8} &52.7 &39.5 &38.3 &34.2 &28.1 &11.1  \\
        & Troika~\cite{huang2024troika} \textsubscript{CVPR'2024} & 49.0 & \underline{53.0} & \underline{39.3} & 22.1 & 66.8 & 73.8 & \underline{54.6} & \underline{41.7} & \underline{41.0} & \underline{35.7} & \underline{29.4} & \underline{12.4}  \\
        \cline{2-14}
        & \textbf{LPR (Ours)} & \textbf{50.6}  & \textbf{53.9} & \textbf{40.0} & \textbf{23.2} & \underline{66.9} & \textbf{76.0} & \textbf{55.7} & \textbf{43.7}   & \textbf{44.1}  & \textbf{39.1}  & \textbf{32.9} & \textbf{14.8} \\
        \hline
        \multirow{10}{*}{\rotatebox{90}{Open-World}} 
        & CLIP~\cite{radford2021clip} \textsubscript{ICML'2021} & 30.1 & 14.3 & 12.8 & 3.0 & 15.7 & 20.6 & 11.2 & 2.2 & 7.5 & 4.6 & 4.0 & 0.3  \\
        & CoOp~\cite{zhou2022CoOp} \textsubscript{IJCV'2022} & 34.6 & 9.3 & 12.3 & 2.8 & 52.1 & 31.5 & 28.9 & 13.2 & 21.0 & 4.6 & 5.5 & 0.7  \\
        & Co-CGE~\cite{mancini2022Co-CGE} \textsubscript{TPAMI'2022} & 38.1 & 20.0 & 17.7 & 5.6 & 59.9 & 56.2 & 45.3 & 28.4 & 33.2 & 3.9 & 5.3 & 0.9  \\
        & CSP~\cite{nayak2022CSP} \textsubscript{ICLR'2023} & 46.3 & 15.7 & 17.4 & 5.7 & 64.1 & 44.1 & 38.9 & 22.7 & 28.7 & 5.2 & 6.9 & 1.2  \\
        & DFSP(t2i)~\cite{lu2023DFSP} \textsubscript{CVPR'2023} & 47.5 & 18.5 & 19.3 & 6.8 & 66.8 & 60.0 & 44.0 & 30.3 & 38.3 & 7.2 & 10.4 & 2.4  \\
        & GIPCOL~\cite{xu2024gipcol} \textsubscript{WACV'2024} & 48.5 & 16.0 & 17.9 & 6.3 & 65.0 & 45.0 & 40.1 & 23.5 & 31.6 & 5.5 & 7.3 & 1.3  \\
        & PLID~\cite{bao2023PLID} \textsubscript{ECCV'2024} & 49.1 & 18.7 & 20.0 & 7.3 & \textbf{67.6} & 55.5 & 46.6 & 30.8 & 39.1 & 7.5 & 10.6 & 2.5  \\
        & CDS-CZSL~\cite{li2024CDS-CZSL} \textsubscript{CVPR'2024} & \underline{49.4} & \underline{21.8} & \underline{22.1} & \underline{8.5} & 64.7 & \underline{61.3} & \underline{48.2} & 32.3 & 37.6 & \underline{8.2} & \underline{11.6} & \underline{2.7}  \\
        & Troika~\cite{huang2024troika} \textsubscript{CVPR'2024} & 48.8 & 18.7 & 20.1 & 7.2 & 66.4 & 61.2 & 47.8 & \underline{33.0} & \underline{40.8} & 7.9 & 10.9 & \underline{2.7}  \\
        \cline{2-14}
        & \textbf{LPR (Ours)} & \textbf{50.1} & \textbf{22.2} & \textbf{22.7} & \textbf{8.9} & \underline{66.9}  & \textbf{61.6} & \textbf{49.8} & \textbf{34.1} & \textbf{44.2}  & \textbf{9.7} & \textbf{12.9} & \textbf{3.7} \\
        \hline
    \end{tabular}
    }
    \label{tab:main_result}
    \vspace{-0.2cm}
\end{table*}
\setlength{\tabcolsep}{6pt}

\subsection{Role of LPR Branches}
LPR employs three distinct branches, denoted as \textit{com}, \textit{sor}, and \textit{osr} (see Fig.~\ref{fig:main_fig}).
The \textit{com} branch functions as a standard composition classifier, while the \textit{sor} and \textit{osr} branches serve as the novel classifiers designed to learn bidirectional relationships between states and objects.
The primary aim is to transform the image feature $\mathbf{x}_i$ within each branch so that it becomes close to the corresponding target text class feature.
To this end, each branch processes $\mathbf{x}_i$ through an Adapter~\cite{gao2024clipadapter}, generating branch-specific features $\mathbf{x}^{c}$, $\mathbf{x}^{s}$, and $\mathbf{x}^{o}$.

In the \textit{com} branch, we compute the cosine similarity between $\mathbf{x}^{c}$ and $\mathbf{T}^c$ to estimate $p(c|\mathbf{x}^c)$ and find the closest composition.
In the \textit{sor} and \textit{osr} branches, directly modeling the bidirectional relationships between states and objects is challenging.
To address this, we adopt a decomposition approach inspired by the \textit{Bayes rule}, where one primitive is conditioned on the probability of the other.
Specifically, in the \textit{sor} branch, we first extract state information and then use it to extract the object information.
To do so, we utilize $\mathbf{T}^s$ as prototypes to extract probabilistic state information from $\mathbf{x}^{s}$ by applying the cross-attention mechanism.
In detail, we project $\mathbf{x}^s$ into the query $\textbf{q}$ and project $\mathbf{T}^s$ into the key $\textbf{K}$ and the value $\textbf{V}$ for the cross-attention.
Intuitively, this process generates a probability distribution that indicates how close $\mathbf{x}^s$ is to each state.
The output of cross-attention, $\bar{\mathbf{x}}^s$, can be interpreted as a state-informed image feature vector.

Next, we utilize $\mathbf{T}^o$ as object prototypes to extract probabilistic object information conditioned on both image and state information.
Here, $\bar{\mathbf{x}}^s$ transforms into the query, while $\mathbf{T}^o$ transforms to serve as both the key and the value.
After passing through the attention and MLP blocks, $\bar{\mathbf{x}}^s$ is transformed into $\bar{\mathbf{x}}^{so}$, a state-conditioned object-informed image feature vector.
Then $\bar{\mathbf{x}}^{so}$ is used to predict the compositional class, represented as $p(c|\bar{\mathbf{x}}^{so})$.

The \textit{osr} branch operates similarly to the \textit{sor} branch, except that it first transforms $\mathbf{x}^o$ into the object-informed image feature vector $\bar{\mathbf{x}}^o$, which is then used to obtain object conditioned state-informed image feature vector, $\bar{\mathbf{x}}^{os}$.
Lastly, $\bar{\mathbf{x}}^{os}$ is utilized to predict the compositional class, expressed as $p(c|\bar{\mathbf{x}}^{os})$.
Please refer to Fig.~\ref{fig:main_fig} for an overview and Fig.~\ref{fig:main_sub} for architectural details.

\subsection{LPR Training and Inference}
During training, we apply one loss to the \textit{com} branch and three losses to the \textit{sor} and \textit{osr} branches:
\begin{equation}
\begin{aligned}
\mathcal{L}_{com} &=  \text{CE}(\mathbf{x}^c, c), \\
\mathcal{L}_{sor} &=  \lambda_1 \text{CE}(\bar{\mathbf{x}}^{so}, c) + \lambda_2\left[ \text{CE}(\mathbf{x}^s, s) + \text{CE}(\bar{\mathbf{x}}^s, o) \right], \\
\mathcal{L}_{osr} &=  \lambda_1 \text{CE}(\bar{\mathbf{x}}^{os}, c) + \lambda_2 \left[ \text{CE}(\mathbf{x}^o, o) + \text{CE}(\bar{\mathbf{x}}^o, s) \right],
\end{aligned}
\end{equation}
where CE refers to the cross-entropy loss function, and $\lambda_1$ and $\lambda_2$ are loss coefficient hyperparameters.
Total loss is calculated as $\mathcal{L} = \mathcal{L}_{com} +  \mathcal{L}_{sor} + \mathcal{L}_{osr}$.
We introduce intermediate cross-entropy losses for $\mathbf{x}^s$, $\bar{\mathbf{x}}^s$, $\mathbf{x}^o$, and $\bar{\mathbf{x}}^o$.
These losses encourage the intermediate features to capture state or object information before applying the cross-attention.

During inference, for the \textit{sor} and \textit{osr} branches, we supplement $p(c|\cdot)$ using $p(s|\cdot)$ and $p(o|\cdot)$.
The predictions from each branch are combined to produce the final probability:
\begin{equation}
\begin{split}
\hat{p}(c \mid \mathbf{x}_i) &= \alpha  p(c \mid \mathbf{x}^c) \\
&+ \frac{\beta}{2}  \Big[p(c \mid \bar{\mathbf{x}}^{so}) +  p(s\mid \mathbf{x}^s)  p(o \mid \bar{\mathbf{x}}^s) \Big]  \\
&+  \frac{\beta}{2}  \Big[ p(c \mid \bar{\mathbf{x}}^{os}) +  p(s \mid \bar{\mathbf{x}}^o)  p(o \mid \mathbf{x}^o) \Big] ,
\end{split}
\end{equation}
where $\alpha$, $\beta$ are scaling hyperparameters.
The composition with the highest probability is determined as the final prediction.

\setlength{\tabcolsep}{11pt}
\begin{table}[tbp]
\centering
\caption{Path ablation results on MIT-States in the open-world setting.}
\vspace{-0.15cm}
\renewcommand{\arraystretch}{1.1}
\resizebox{1.0\columnwidth}{!}{
\begin{tabular}{ccc|cccc}
\hline
\multicolumn{3}{c|}{Branch} & \multicolumn{4}{c}{MIT-States} \\
\hline
$\textit{com}$ & $\textit{sor}$ & $\textit{osr}$ & S & U & HM & AUC \\
\hline
\checkmark &  &  & 47.5 & 16.1 & 17.6 & 5.9 \\ 
 & \checkmark &  & 49.4 & 22.3 & 22.2 & 8.6 \\  
 &  & \checkmark & 49.6 & 22.3 & 22.4  & 8.7\\  
\checkmark & \checkmark &  & 49.8 & 22.1 & 22.6 & 8.8 \\
\checkmark &  & \checkmark & 50.1 & 22.1 & 22.5  & 8.8 \\
 & \checkmark & \checkmark &  49.7 & 22.4 & 22.3  & 8.7 \\
\checkmark & \checkmark & \checkmark &  50.1 & 22.2 & 22.7  & 8.9 \\
\hline
\end{tabular}
}
\label{tab:path_ablation}
\vspace{-0.2cm}
\end{table}
\setlength{\tabcolsep}{6pt}

\section{Experimental Results}

\subsection{Setup}
We evaluate the performance of our model using three CZSL benchmark datasets. 
MIT-States~\cite{mitstates} consists of 115 states, 245 objects, and 1,962 compositions.
UT Zappos~\cite{utzappos} contains 16 states, 12 objects, and 116 compositions.
C-GQA~\cite{cgqa} includes 413 states and 674 objects, with over 7,000 compositions.

During inference, two different evaluation settings are used: \textit{closed-world} and \textit{open-world}.
In the closed-world setting, the model is tested on a constrained set of feasible compositions, including both seen and unseen compositions, as in standard generalized zero-shot learning~\cite{pourpanah2022review, kim2022semantic, kim2023vision}.
In the open-world setting, the model is evaluated on the entire compositional space \(\mathcal{C}\), which includes all possible combinations of states and objects~\cite{ mancini2021open}.
Therefore, an open-world setting is more challenging because the model should not only predict unseen compositions but also filter out infeasible compositions.

LPR is implemented using a pre-trained CLIP ViT-L/14 model and is trained and evaluated on a single NVIDIA A100 GPU.
Three Adapters are implemented following CLIP-Adapter~\cite{gao2024clipadapter}.
The hyperparameter $\alpha$ is set to 0.4 (MIT-States), 0.7 (UT-Zappos), and 0.4 (C-GQA).
The hyperparameters $\lambda_1$ and $\lambda_2$ are set to 2.0 and 1.5 for both MIT-States and C-GQA, and 3.0 and 1.0 for UT-Zappos.

\pgfplotsset{compat=1.18} 
 
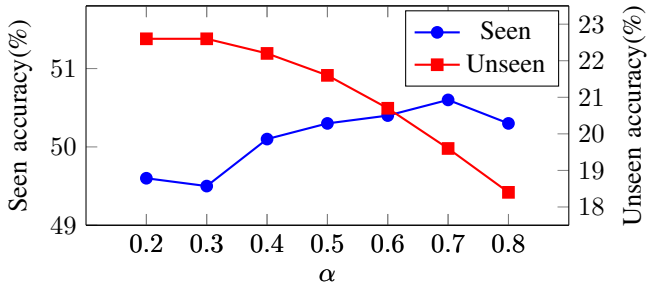
\begin{figure}[t]
    \centering
\begin{tikzpicture} 
\begin{axis}[
    width=8.0cm,
    height=4.5cm,  
    axis y line*=left,
    xlabel=$\alpha$,
    xmin=0.1, xmax=0.9,
    ylabel=Seen accuracy(\%),
    ymin=49.0, ymax=51.8,
    xtick={0.2,0.3,0.4,0.5,0.6,0.7,0.8},
    ytick={49,50,51},
    ] 
    
\addplot[
  color=blue, 
  thick, 
  mark=*] 
  coordinates {(0.2,49.6) (0.3,49.5) (0.4,50.1) (0.5,50.3) (0.6,50.4) (0.7,50.6) (0.8,50.3) };\label{plot_one} 

\end{axis}

\begin{axis}[
    width=8.0cm,   
    height=4.5cm,   
    axis y line*=right,
    xlabel=$\alpha$,
    xmin=0.1, xmax=0.9, 
    xtick={0.2,0.3,0.4,0.5,0.6,0.7,0.8}, 
    ylabel={Unseen accuracy(\%)},
    ymin=17.5, ymax=23.5,
    ytick={18,19,20,21,22,23},
    ]
\addlegendimage{/pgfplots/refstyle=plot_one}\addlegendentry{Seen} 
\addplot[
  color=red, 
  thick, 
  mark=square*] 
  coordinates {(0.2,22.6) (0.3,22.6) (0.4,22.2) (0.5,21.6) (0.6,20.7) (0.7,19.6) (0.8,18.4)};
\addlegendentry{Unseen}
\end{axis} 
\end{tikzpicture}
\vspace{-0.4cm}
\caption{Effect of $\alpha$ during inference on MIT-States in the open-world setting.}
\vspace{-0.1cm}
\label{fig:inference_weight_ablation} 
\end{figure}

\subsection{CZSL Performance}
Table~\ref{tab:main_result} presents the performance of the LPR and previous CZSL techniques.
The results show that LPR significantly outperforms all existing methods across all datasets.
Specifically, on the C-GQA dataset, LPR achieves a harmonic mean accuracy (HM) of 32.9\% and 12.9\% for closed-world and open-world settings, representing 3.5 and 2.0 percentage points improvement over previous SOTA~\cite{huang2024troika}.

\subsection{Ablation Study}
To demonstrate the effectiveness of utilizing three branches (\textit{com}, \textit{sor}, and \textit{osr}) in the proposed architecture, we conduct a path ablation study in Table~\ref{tab:path_ablation}.
The results show that the highest HM is achieved when all three branches are activated.
Moreover, we can observe that combining the \textit{com} branch with other branches consistently leads to improved performance.
This implies that the newly proposed branches equip the ability to recognize unseen compositions that the \textit{com} branch fails.

In addition, we investigate the effect of the weight hyperparameter $\alpha$ in Fig.~\ref{fig:inference_weight_ablation}.
As expected, with larger $\alpha$, the influence of the \textit{com} branch increases, leading to a gradual improvement in seen accuracy. 
On the opposite, the unseen accuracy increases as $\alpha$ decreases.
We empirically find the optimal balance between $\alpha\in [0.2, 0.4]$.

\begin{figure}[t]
    \centering
    \includegraphics[width=0.95\columnwidth]{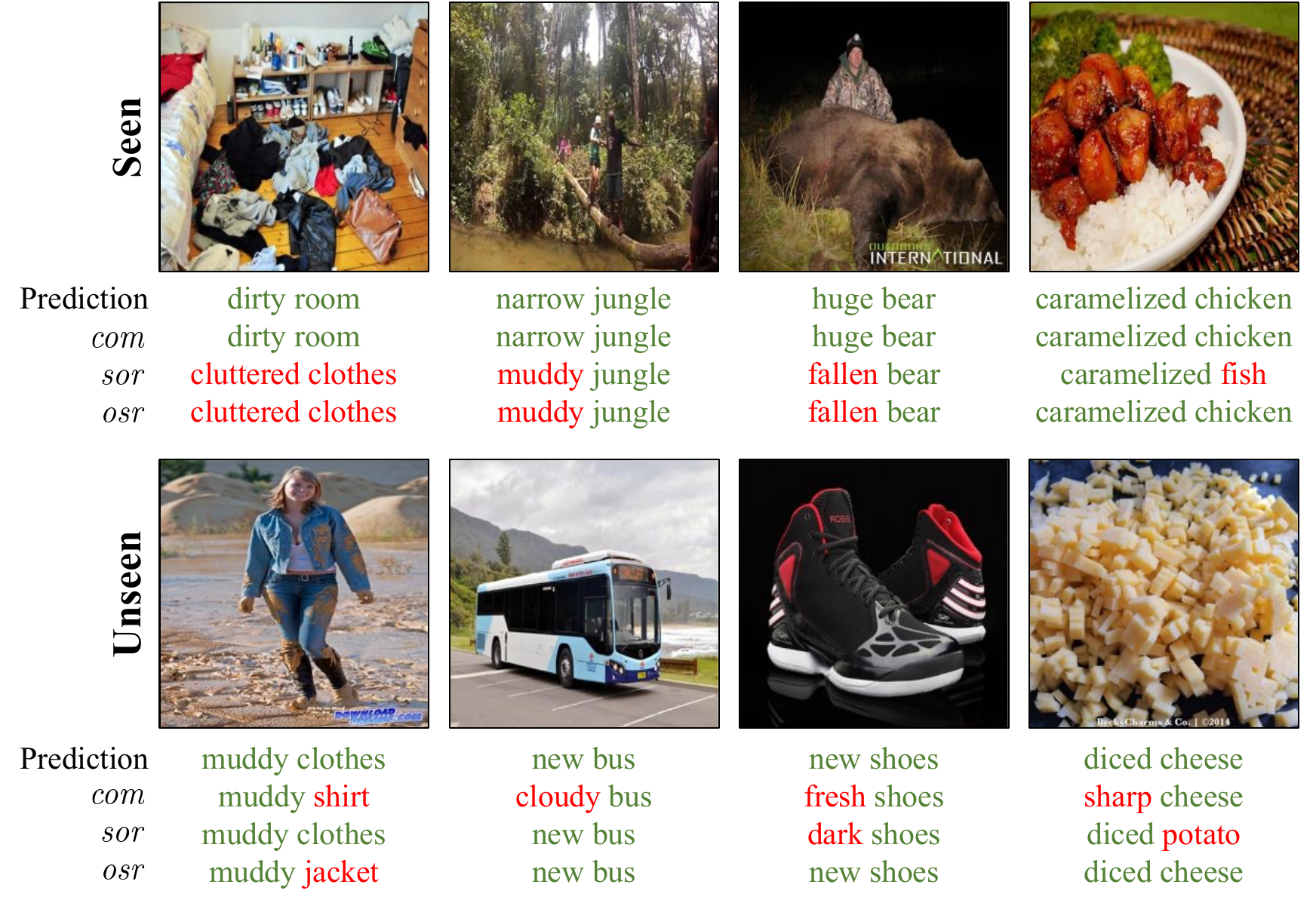}
    \vspace{-0.25cm}
    \caption{Composition classification results from the three different branches. `Prediction' indicates the final results. Green and red words indicate correct and incorrect predictions, respectively.}
    \vspace{-0.25cm}
    \label{fig:qualitative}
\end{figure}

\subsection{Sensitivity to Hyperparameters}
We conduct additional experiments to test the hyperparameter sensitivity in Table~\ref{tab:ut_zappos_inference_weight_ablation}.
As noted in Sec.~\uppercase\expandafter{\romannumeral3}-A, $\alpha$ and $\beta$ are shared between MIT-States and C-GQA datasets.
To assess the generalizability of our hyperparameters, we evaluate their configuration to the UT-Zappos dataset (denoted as \textbf{default}).
We denote LPR in the setup from Table~\ref{tab:main_result} as \textbf{best}.

While there exists a slight degradation in performance, LPR consistently outperforms Troika~\cite{huang2024troika} in both closed and open-world setups, indicating that our model shows robust performance with minimal hyperparameter tuning.

\setlength{\tabcolsep}{12pt}
\begin{table}[t]
    \centering
        \caption{Effect of hyperparameters on UT-Zappos dataset.
        The best results are in \textbf{bold}.
        The second-best results are \underline{underlined}.
        }
        \vspace{-0.2cm}
    \renewcommand{\arraystretch}{1.1}
    \resizebox{1.0\columnwidth}{!}{
    \begin{tabular}{c|c|c c}
        \hline
        \multirow{2}{*}{Setup} & \multirow{2}{*}{Method} & \multicolumn{2}{c}{UT-Zappos} \\
        & & HM & AUC \\
        \hline
        \multirow{3}{*}{Closed-World}
        & Troika & 54.6 & 41.7 \\
        & \textbf{LPR ($\alpha = 0.4$, default)} & \textbf{55.7} & \underline{43.2} \\
        & \textbf{LPR ($\alpha = 0.7$, best)} & \textbf{55.7} & \textbf{43.7} \\
        \hline
        
        \multirow{3}{*}{Open-World} 
        & Troika & 47.8 & 33.0\\  
        & \textbf{LPR ($\alpha = 0.4$, default)} & \underline{48.9} & \underline{33.6} \\
        & \textbf{LPR ($\alpha = 0.7$, best)} & \textbf{49.8} & \textbf{34.1} \\
        \hline
    \end{tabular}
    }
    \label{tab:ut_zappos_inference_weight_ablation}
    \vspace{-0.3cm}
\end{table}
\setlength{\tabcolsep}{6pt}

\subsection{Qualitative Analysis}
We visualize the qualitative examples in Fig.~\ref{fig:qualitative}.
While all compositions are correctly classified, each branch shows slightly different predictions.
For the seen class samples (upper row), the \textit{com} branch is more accurate than the other two branches. 
However, the \textit{com} branch sometimes predicts awkward compositions for unseen images (lower row), demonstrating the vulnerability of using the \textit{com} branch alone for the CZSL task.
We note that \textit{sor} and \textit{osr} branches usually generate reasonable predictions regardless of the exact match for the given composition class label.

\section{Related Work}

Conventional CZSL approaches~\cite{mancini2022Co-CGE, karthik2022kg, li2022siamese, li2020symmetry} utilize two classifiers to independently identify states and objects. 
Each classifier predicts either the state or the object, and the predicted state and object are combined to determine the compositional class. 
In recent approaches~\cite{huang2024troika,li2024CDS-CZSL, wang2023HPL}, trainable state and object tokens are employed to generate textual prompts for compositions, fully utilizing VLMs' ability to map images and text into a shared feature space~\cite{kim-etal-2024-preserving}.

Our work is significantly distinct from previous works because the model systematically considers the relationships between states and objects.
Key distinguishing feature of LPR compared to previous works is the decomposition of joint probabilities using two branches, where cross-attention-based architecture captures the relationships between primitives.

\section{Conclusion}
In this paper, we proposed a novel CSZL framework, LPR, which implicitly learns the relationships between states and objects as a probabilistic model.
LPR introduces two new prediction branches that utilize the cross-attention mechanism between image and text feature embeddings in two different orders: state-to-object and object-to-state.
Experiments showed that LPR achieves the best CZSL accuracy, especially for unseen class classification performance.

\newpage
\bibliographystyle{IEEEbib}
\bibliography{reference}

\end{document}